# Data-Driven Meets Navigation: Concepts, Models, and Experimental Validation

Itzik Klein
The Autonomous Navigation and Sensor Fusion Lab
Hatter Department of Marine Technologies
University of Haifa,
Haifa, Israel

**Abstract**

The purpose of navigation is to determine the position, velocity, and orientation of manned and autonomous platforms, humans, and animals. Obtaining accurate navigation commonly requires fusion between several sensors, such as inertial sensors and global navigation satellite systems, in a model-based, nonlinear estimation framework. Recently, data-driven approaches applied in various fields show state-of-the-art performance, compared to model-based methods. In this paper we review multidisciplinary, data-driven based navigation algorithms developed and experimentally proven at the Autonomous Navigation and Sensor Fusion Lab (ANSFL) including algorithms suitable for human and animal applications, varied autonomous platforms, and multi-purpose navigation and fusion approaches.

## 1. Introduction

Navigation is the science and art of estimating the position, velocity, and orientation of a platform. One of the means to perform navigation is using a dead reckoning (DR) approach. In DR, given initial conditions, velocity or acceleration measurements are integrated to obtain the position. An inertial navigation system (INS) is the most popular tool working with DR principles. Its popularity stems from these facts: it provides a full navigation solution (position, velocity, and orientation), it is a standalone system capable of working in any environment (land, air, underground, underwater, indoors), and it is available in many different grades (ranging from low-cost low-performance to high-cost high-performance systems) [1-3].

The INS contains two types of three-orthogonal inertial sensor sets: accelerometers to measure the specific force vector and gyroscopes to measure the angular velocity vector of the platform. The inertial measurements are plugged into the INS equation of motion to estimate the INS navigation solution. As the inertial sensor measurements contain noise and other error sources, solution drift occurs upon integration when calculating the navigation solution [4-5].

Consequently, regardless of the inertial sensors' quality, the navigation solution drifts with time. To circumvent the INS navigation solution drift, the INS is fused with external sensors or information [6]. Due to the nonlinear nature of the INS equations of motion, the fusion process is carried out using a nonlinear filter by frequently applying an extended Kalman filter (EKF) [7-8]. External sensors include, for example, a global navigation satellite systems (GNSS) receiver to provide position and/or velocity updates for land and airborne platforms; while underwater, a Doppler velocity log is used to estimate the platform velocity vector [9].

All the above algorithms—namely the inertial navigation equations of motion and the navigation filter—operate in a model-based fashion. In the past 20 years the artificial intelligence (AI) field has led the fourth industrial revolution and presents an exciting and blooming research field. AI is expected to automate routine labor, understand images and speech, make diagnoses in medicine, and support basic scientific research [10]. A machine learning (ML) algorithm is an approach to AI that is capable of learning from data. A more formal definition of ML is this [11]: A computer program is said to learn from experience E with respect to some class of tasks T and performance measure P, if its performance at tasks in T, as measured by P, improves with experience E.

Deep learning (DL) is an ML set of algorithms that represents the world as a nested hierarchy of concepts, with each concept defined in relation to simpler concepts, and abstract representations computed in terms of less abstract ones [10]. In practice, such a hierarchy is represented by layers and the number of units in each layer. The most commonly applied neural network is known as convolutional neural networks (CNN) [12,13] and is used for processing data with a grid-like topology such as time series data (e.g., inertial sensor readings). As implied by its name, CNN employs a convolution operator. CNN has proven its ability in various domains such as image processing and natural language processing.

To apply any ML or DL algorithm, a set of training examples known as the train dataset is employed. As both ML and DL approaches learn from data, they are also referred to as data-driven approaches.

In this paper we review multidisciplinary data-driven based navigation algorithms developed and experimentally proven at the Autonomous Navigation and Sensor Fusion Lab (ANSFL) at the University of Haifa. The purpose of those algorithms is to enhance common navigation and estimation tasks and open new possibilities for accurate and robust inertial navigation and fusion. Data-driven inertial navigation topics included in this paper highlight hybrid learning and end-to-end learning approaches for different platforms and applications. They include autonomous underwater vehicle (AUV) navigation, pure inertial navigation for quadrotors, pedestrian dead reckoning (PDR) using a smartphone, attitude and heading reference systems, land vehicle navigation, identifying illness in dogs, and multi-platform algorithms such as denoising.

The rest of the paper is organized as follows: Section 2 provides ANSFL inertial data-driven research in a variety of applications. Section 3 focuses on autonomous platform navigation while Section 4 does the same for smartphones. Section 5 addresses self-driving and off-road vehicle navigation, and Section 6 provides learning approaches suitable for any type of platform. Section 7 describes the usage of inertial sensors to identify ataxia in dogs. Finally, Section 8 gives the conclusions.

## 2. ANSFL Inertial Data-Driven Research

At ANSFL we derived novel, inertial, data-driven approaches, including hybrid and end-to-end methods in various navigation domains. They are summarized in Table 1 using a platform type categorization. The table presents the research goal, the data-driven approach used in each research, the experimental validation, and the paper citation.

Table 1: ANSFL inertial data-driven research categorized by platform type

|   | Platform Type | Research Goal | Data-Driven Approach | Experimental Validation | Ref |
|---|---|---|---|---|---|
| 1 | Autonomous underwater vehicle | Improve accuracy and robustness of the Doppler velocity log estimated velocity vector | End-to-end deep learning | Snapir AUV | [14] |
|   |   | Cope with partial Doppler velocity log measurements using Doppler velocity log and/or inertial data | End-to-end deep learning | Snapir AUV | [15, 16] |
| 2 | Quadrotor | Mitigate pure inertial solution drift | End-to-end deep learning | Matrice 300 Quadrotor | [17] |
| 3 | Smartphone | Learn step length estimation in a model-based pedestrian dead reckoning framework | Hybrid approach | Multiple users with multiple smartphones | [18] |
|   |   | Improve pedestrian dead reckoning accuracy using deep learning tools | End-to-end deep learning | Multiple users with multiple smartphones | [19] |
|   |   | Estimate walking direction for enhanced pedestrian dead reckoning | Hybrid approach | Single user single smartphone | [20] |
|   |   | Recognize human activity and smartphone location for improved pedestrian dead reckoning performance | Deep learning, transformers, transfer learning | Multiple users with multiple smartphones | [21-24] |
|   |   | Measure inertial sensors' sensitivity to home appliances | Machine learning | Smartphone and home appliances | [25] |
| 4 | Self-Driving Cars | Learn vehicle trajectory under uncertainty | Hybrid approach | Land vehicles | [26] |
| 5 | Off-road vehicles | Evaluate Dozer grading policy under sensor uncertainty | Hybrid approach | Dozer prototype | [27] |

|   | Platform Type | Research Goal | Data-Driven Approach | Experimental Validation | Ref |
|---|---|---|---|---|---|
| 6 | Any-Platform Basic Inertial Tasks | Accelerometer denoising using learning algorithms | End-to-end machine learning | Low and high grade inertial sensors | [28] |
|   |   | Zero order gyroscope calibration using learning algorithms | End-to-end deep learning | Low and high grades inertial sensors | [29] |
|   |   | Estimate feasibility of stationary coarse alignment using learning approaches | Machine learning | Low grade inertial sensors | [30, 31] |
|   | Any-Platform Attitude and Heading Reference Systems | Estimate attitude in presence of accelerations for a walking user | Data-driven optimization | Multiple users with multiple smartphones | [32] |
|   |   | Derive an attitude and heading reference system based on a data-driven approach | Data-driven optimization | Multiple users with multiple smartphones | [33] |
|   |   | Estimate enhanced attitude in the presence of accelerations for a walking user | Hybrid approach | Multiple users with multiple smartphones | [34] |
|   | Any-Platform Navigation Filter | Learn the optimal inertial time step in a navigation filter | Hybrid approach | Matrice 300 Quadrotor | [35] |
|   |   | Estimate adaptive process noise covariance using deep learning | Hybrid approach | Matrice 300 Quadrotor | [36] |
| 7 | Dogs | Classify ataxia using inertial sensors | End-to-end machine learning | Multiple dogs with low grade inertial sensors | [37] |

In the following sections we describe ANSFL's research in all fields as presented in Table 1 including six unique dedicated algorithms for specific platforms and algorithms that can be implemented in any type of platform, for a total of 24 novel learning algorithms. All the figures in this paper were generated with the same color map, where each color represents the same block type. Model-based blocks are colored gray, AI blocks are orange, the measurement block is light blue, and the output of the downstream algorithm is blue.

# 3. Autonomous Platforms Navigation

## 3.1 Autonomous Underwater Vehicles

Covering about two-thirds of the earth's surface, oceans have a great impact on mankind, both now and for the future. Advances in many technological fields led the development of autonomous underwater vehicles (AUV), which are used for scientific, industrial, military, and search and rescue applications. One of the most commonly used AUV navigation sensor is the Doppler velocity log (DVL), which transmits four acoustic beams. When they are received, the AUV velocity vector can be estimated using a model-based approach, usually a parameter estimation method such as least-squares. Once the velocity vector is obtained, it can be integrated to obtain the AUV position or fused with inertial sensors in a navigation filter.

To improve the accuracy of velocity estimation using a model-based approach, in [14] we proposed an end-to-end network, BeamsNet. Our motivation stemmed from well-known deep learning abilities such as an inherit noise reduction capability and the ability to capture non-linear behavior and other uncertainties in the input data. BeamsNet has two different versions as shown in Figure 1. One of them employs the current DVL beam measurements and inertial sensor readings, while the other leverages only from the DVL measurements, utilizing past and current DVL measurements for the regression process.

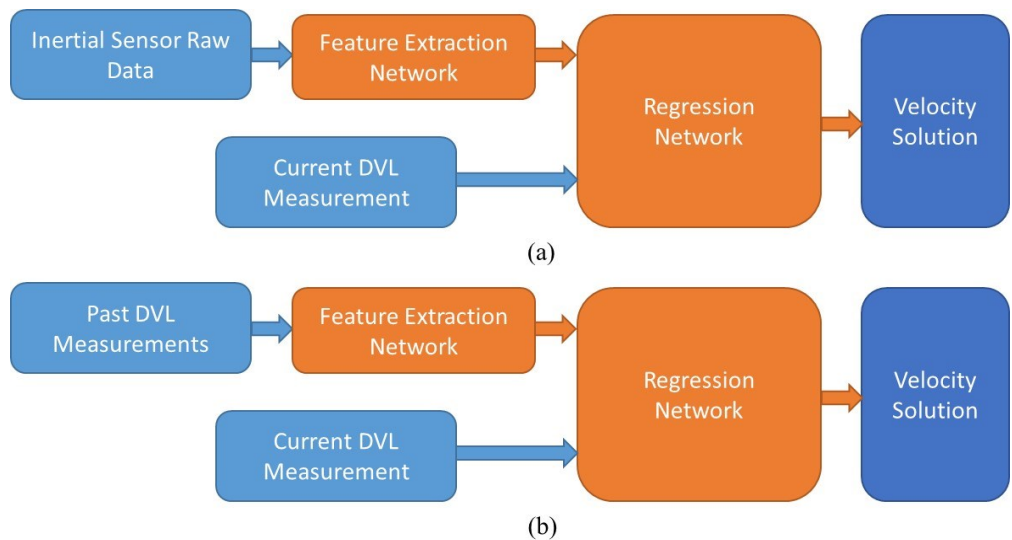

Figure 1. Two BeamsNet architectures: (a) both inertial data and current DVL (b) past and current DVL measurements, are plugged into a regression network to regress the velocity vector

Sea experiments using the University of Haifa's Snapir AUV (ECA Robotics A18D) were done in the Mediterranean Sea. Figure 2 shows Snapir AUV seconds before starting its underwater mission. More than four and a half hours of data was gathered from the AUV's inertial sensors and DVL readings. With respect to the root mean square error, utilizing this dataset, we showed that both BeamsNet architectures improve the model-based approach by 60%.

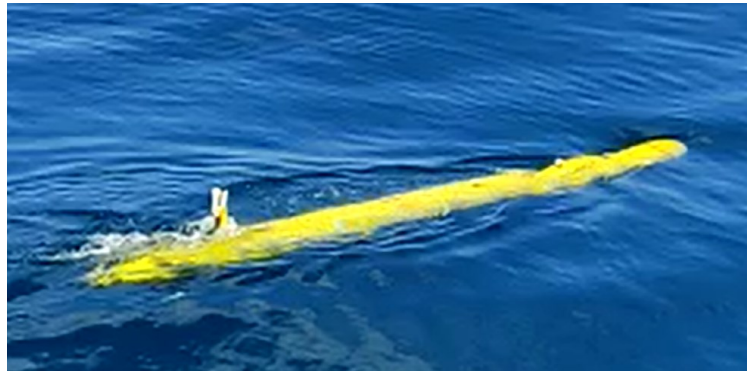

Figure 2. Snapir AUV, a 5.5m long ECA Robotics A18D, seconds before starting its underwater mission

However, some of the DVL beams may experience outages (partial beam availability) when the AUV operates in complex environments such as extreme roll/pitch angles, passing over fish and other sea creatures, sound scattering, and passing over trenches in the seafloor. In such scenarios, where less than three beams are available, the DVL cannot estimate the AUV velocity vector and hence a drift in the position estimation occurs. To circumvent such situations, we offered an end-to-end network capable of regressing the missing beams using past DVL measurements [15]. Snapir AUV was used here also, with a smaller dataset, to show that our network is capable of regressing the missing beams' velocity and thereby estimating the AUV velocity. Recently, in [16] we modified both BeamsNet architectures to handle missing beam scenarios using the same dataset as in [14] and showed better performance than in [15] when estimating the missing beam velocities.

*3.2 Quadrotor Dead Reckoning*

Quadrotors use an inertial navigation system combined with a global navigation satellite systems receiver for outdoor applications and a camera for indoor/outdoor applications. For various reasons, such as lighting conditions or satellite signal blocking, the quadrotor's navigation solution depends only on the inertial navigation system solution. As a consequence, the navigation solution drifts in time due to errors and noises in the inertial sensor measurements. To handle such situations and bind the solution drift, we developed QuadNet, a hybrid framework for quadrotor dead reckoning to estimate the quadrotor's three-dimensional position vector at any user-defined time rate [17].

The requirement for QuadNet is that the quadrotor is flown in a periodic motion trajectory instead of a straight-line trajectory. In this manner, the inertial sensor readings vary during the motion and the repetitive motion helps the network to accurately perform its task. As a hybrid approach, QuadNet uses both neural networks and model-based equations. Supervised regression networks are employed to regress the change in distance and height, and model-based equations are used to estimate the orientation using attitude and heading reference system (AHRS) algorithms during the quadrotor operation. QuadNet only requires the inertial sensor readings to provide the position vector, as illustrated in Figure 1.

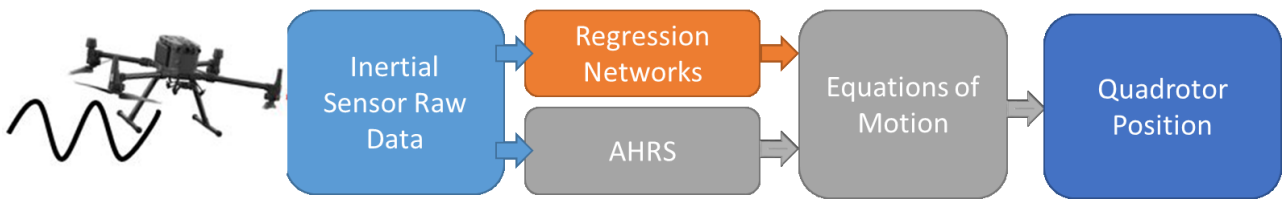

Figure 3. QuadNet hybrid structure for estimating the quadrotor position in situations of pure inertial navigation

Experimental results with the DJI Matrice 300 quadrotor—shown in Figure 4—and two mounted inertial measurement units (IMU) are provided to show the benefits of using the proposed approach. In model-based pure inertial navigation, the root mean square error (RMSE) when estimating the distance is between 40 and 120 meters while QuadNet manages to reduce the error to less than four meters. In the same manner, the RMSE error in the estimated quadrotor height was between 20 and 80 meters while QuadNet manages to reduce the error to less than two meters.

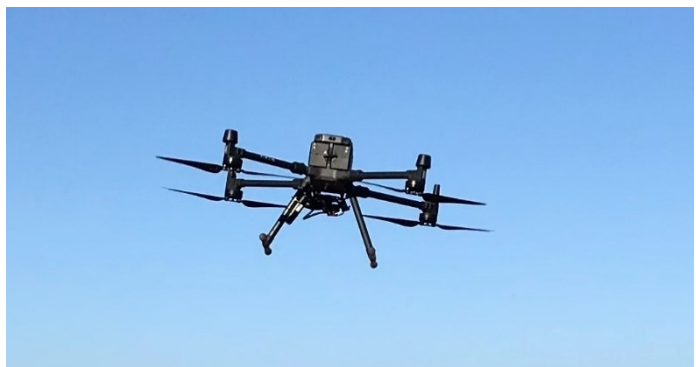

Figure 4. Matrice 300 quadrotor

## 4. Smartphone Navigation

*4.1 Pedestrian Dead Reckoning*

Pedestrian dead reckoning (PDR) algorithms were found to be very sensitive to pedestrian activity (e.g., walking or running) and smartphone location (e.g., in pocket or swinging). This led to the initial implementation of supervised machine learning classification algorithms for human activity recognition and smartphone location recognition. By knowing the activity, appropriate gain values are selected, and as a consequence the algorithm drift is mitigated [38]. Thus, a prior task of motion recognition is added to the model-based PDR algorithm.

Our work [18] proposed StepNet, a family of deep learning based PDR approaches. Three different approaches for the regression task were proposed and examined: 1) step length, 2) change in distance, and 3) PDR gain to be used in the model-based PDR. Our results showed that regressing the change in distance, in any required time interval, is the best choice for regressing the distance. Later, our PDRNet framework [19] showed superior performance over RoNIN and model-based PDR approaches while using only the accelerometer and gyroscope readings. A residual network (ResNet) was employed to map from inertial sensor measurements to changes in distance and heading. The main contribution of our proposed approach is the definition of a modular structure framework to perform PDR using deep learning. Specifically, the basic building blocks that should be applied in any similar approach were defined and are presented in Figure 5. They include the smartphone location recognition network, ResNet blocks in the regression model, and definition of the input/output of the network. In addition, we show a strong connection between the change in distance and heading estimates. This leads to the conclusion that future research in the field should focus on the combined estimation instead of splitting the two.

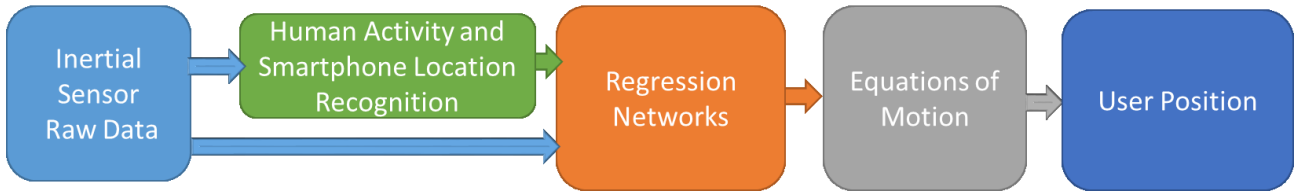

Figure 5. Hybrid PDR block-diagram using regression networks to estimate the change of distance and heading to estimate the user position

One of the major challenges in pedestrian dead reckoning methods is estimating the misalignment angle between the user's forward axis and the smartphone's frame to find the user's heading direction, a problem known as walking direction estimation. Recently, we proposed a novel deep neural network architecture for extracting the motion vector in the smartphone coordinates using accelerometer measurements [20]. To that end, temporal convolutions and multiscale attention layers were used and a unique rotation-invariant property was analytically derived. The proposed framework integrates our network with a geometric calculation, using the gravity and geomagnetic directions, to convert the motion vector into the heading angle.

*4.2 Human Activity and Smartphone Location Recognition*

As we explain Section 4.1, knowledge of the human activity (walking, running, etc.) and the smartphone location on the user (in pocket, swinging, etc.) is needed to improve model-based and learning based PDR using the smartphone inertial sensors. Our initial work in the field addressed the smartphone location recognition (SLR) classification task [21]. We defined the basic building blocks required to perform SLR using neural networks, as shown in Figure 6. Generally, the accelerometer and gyroscope readings are inserted into a classification network to yield the SLR result. Our paper addressed four different smartphone locations: 1) texting, 2) swinging, 3) talking, and 4) in pocket. We evaluated the approach on 31 hours of recorded data collected from 107 people appearing in eight different datasets.

In such a supervised setup, if the smartphone is in an unidentified location (that is not part of the training set), the classifier is forced to identify it as one of the existing modes on which it was trained. Such classification errors degrade the navigation solution accuracy. Therefore, also in [21], a feasibility study to cope with unknown smartphone locations used only the accelerometer readings and a binary network. Relying on this work, in [22] we proposed two end-to-end, learning-based approaches for unknown mode recognition: (1) a supervised learning method that requires labelling the known modes during the training phase, and (2) an unsupervised approach without the requirement to label training data. In both approaches, a feature representation space is extracted and fed into a K-nearest neighbor algorithm to detect unknown modes. Using six different datasets, we showed that our approaches achieve more than 93% accuracy in identifying the unknown modes. The proposed approaches can be easily applied to any other classification problems containing unknown modes.

To further improve the SLR and human activity recognition (HAR) we explored the possibility of using transformers for inertial data classification [23]. Several datasets, with more than 27 hours of inertial data readings collected by 91 users were examined. Our proposed approach consistently achieves better accuracy and generalizes better across all examined scenarios and datasets. Later, we proposed a novel, time series encoding approach, from accelerometer and gyroscope readings to inertial images (INIM), and also demonstrated transfer learning from the computer vision domain (using ImageNet) to the inertial based classification of SLR and HAR tasks [24].

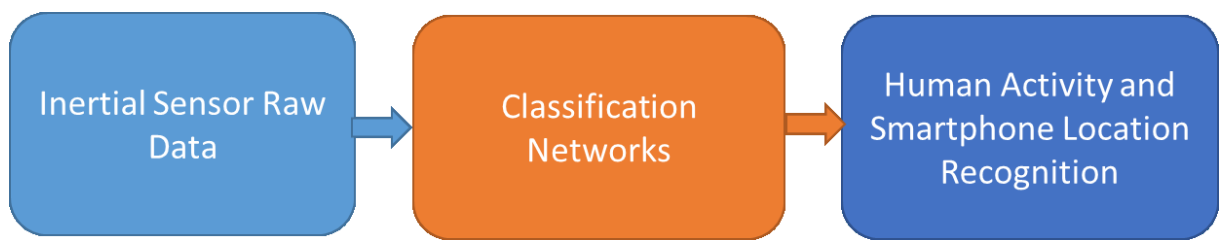

Figure 6. A general block diagram for human activity and smartphone location recognition using neural networks and inertial sensors

*4.3 Anomaly Detection in Inertial Sensors*

It is well known that magnetometers suffer from magnetic interference and as a consequence produce erroneous measurements. In [25] we present our preliminary work demonstrating that accelerometers and gyroscopes are prone to interference from home appliances. We employ the multivariate Gaussian distribution anomaly detection (MGAD) model as a semi-supervised learning method because it can perform anomaly detection for multivariate time series data. Once an anomaly is detected in the inertial measurement, we use a heuristic approach to ignore it and replace the measurement, so the downstream task working with the inertial data is not influenced. A block diagram of this approach is shown in Figure 7.

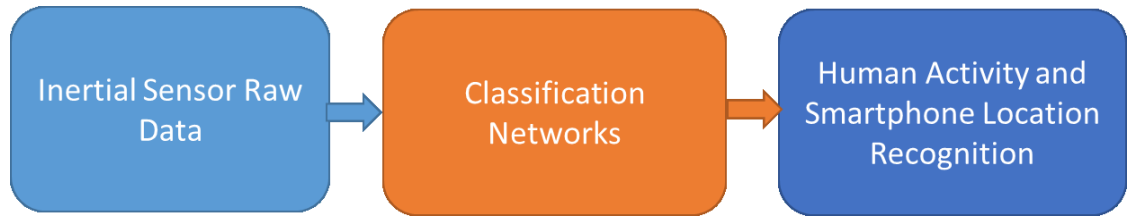

Figure 7. Block diagram of the proposed approach for inertial sensor anomaly detection and rejection given inertial sensor readings and any downstream task

## 5. Self-Driving and Off-Road Vehicle Navigation

Vehicle tracking is one of the tasks required for self-driving cars. A very common and easy way to perform tracking is using GNSS signals in a linear Kalman filter framework. Such a filter requires knowledge of the expected vehicle dynamics and statistics of the system and measurement models. Yet, prior assumptions made while determining these models do not hold in real-life scenarios. As a consequence, the overall filter performance degrades and in some situations the estimated states diverge. One of the common model-based approaches to overcome the uncertainty in vehicle kinematic trajectory modeling is to add additional artificial process noise, which, in some situations (when the assumptions hold), damages the filter's performance. To cope with such situations, we proposed a two-stage hybrid adaptive Kalman filter framework [26]. A bi-directional, long, short-term memory, recurrent neural network is used to regress the vehicle's geometrical feature. Combined with other features, those are used as input to a supervised learning model, providing the measure of uncertainty, which is the actual process noise covariance. In turn, this process noise is used in the Kalman filter to estimate the vehicle position as illustrated in Figure 8.

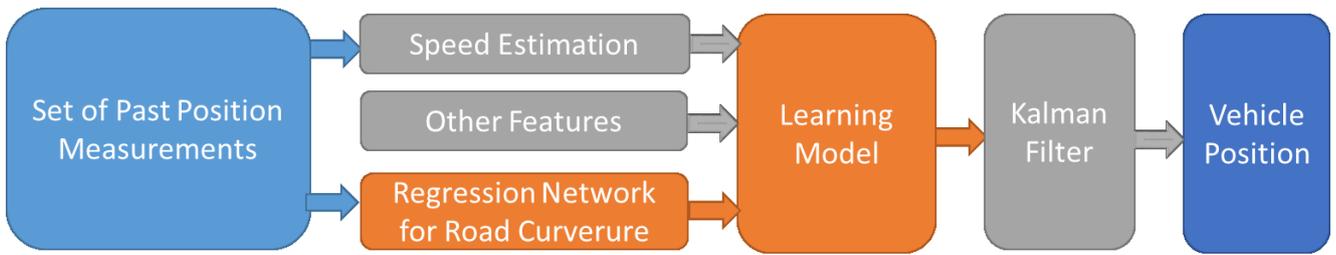

Figure 8. A two-stage hybrid learning model for regressing vehicle uncertainty required for vehicle position estimation via a linear Kalman filter

In addition to the extensive research aimed at self-driving cars, recently, more attention is given to autonomous off-road vehicles and in particular to those operating in a construction site. There, one of the important tasks is surface grading—a process of leveling an uneven area containing pre-dumped sand piles. This task is often carried out by a dozer, a key machinery tool at any construction site. Current attempts to automate surface grading assume perfect localization, an assumption not valid in real world scenarios, which may lead to degraded performance and sometimes to mission failure.

In [27] we first demonstrate the hypothesis that an agent trained to perform a grading task assuming perfect localization conditions, underperforms when processing real-world uncertainties present in its sensor readings. To tackle such situations, we proposed a training regime that takes into account the sensory noise while training the agent, and as a consequence produces a robust policy in the presence of uncertainties. Multiple evaluations and experiments showed that our training procedure indeed improves the dozer performance by 40%.

**6 Any-Platform Algorithms**

*6.1 Basic Inertial Tasks*

When working with inertial sensors, regardless of the application, several operations to ensure effective operation of those sensors are applied, including inertial sensor denoising and calibration. Further, when working with inertial sensor readings in an inertial navigation system, sometimes the initial attitude is determined by the inertial sensor readings in a process known as self-coarse alignment. All three tasks use model-based approaches.

Inertial measurements are denoised by conventional model-based denoising techniques. In some situations, the effectiveness of these methods becomes bounded by an uncompensated error, and consequently, most of the noise present in the measurements is not reduced. To mitigate such common situations we proposed a set of data-driven models that successfully managed to denoise accelerometer signals, dramatically improving not only its signal-to-noise ratio but also subsequent navigation tasks [28]. Our experimental results showed a clear advantage of the learning methods over model-based signal processing algorithms, presumably due to their generalization ability to compensate for a wide range of error sources.

While denoising addresses the random noise part in the inertial measurement, calibration aims to eliminate the deterministic part. Usually, the inertial sensors are calibrated in the factory during manufacture. Upon leaving the factory, the calibrated sensor starts accumulating error sources that slowly degrade its precision and reliability. Therefore, periodic recalibration is required to restore intrinsic parameters. In [29] we show how bias elimination in low-cost gyroscopes is considerably faster using a unique convolutional neural network framework. As illustrated in Figure 9, the raw gyroscope readings are plugged into a bias regression network to estimate the gyro bias. The strict constraints of model-based approaches are swapped for a learning-based regression model that reduces the time-consuming average time, exhibiting efficient sifting of noise from the actual bias.

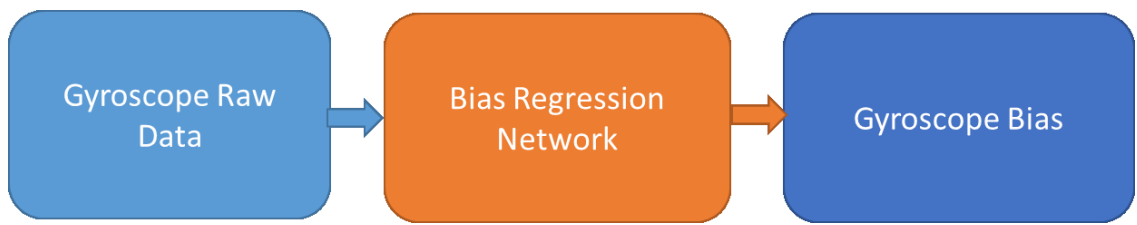

Figure 9. Gyroscope bias regression structure

In our work [30], the feasibility of using machine learning algorithms for stationary coarse attitude alignment was examined and showed potential by improving the commonly used model-based approaches. Later in [31], we derived a machine learning coarse alignment (MLCA) framework including a pyramidal methodology approach to cope with computing hardware limitations. Using raw accelerometer data, features are extracted and introduced into a learning algorithm to regress the initial attitude, as presented in Figure 10. MLCA elaborates upon our work [30] with broader research scenarios, methodology, and field experimental results. Our results show an improvement of more than 70% compared to model-based approaches. Also, the overall time required in MLCA to obtain the same accuracy is lower than the model-based approach. Improved accuracy and required time for the alignment task are particularly critical for platforms operating in pure inertial navigation and when time constraints are imposed on the alignment time.

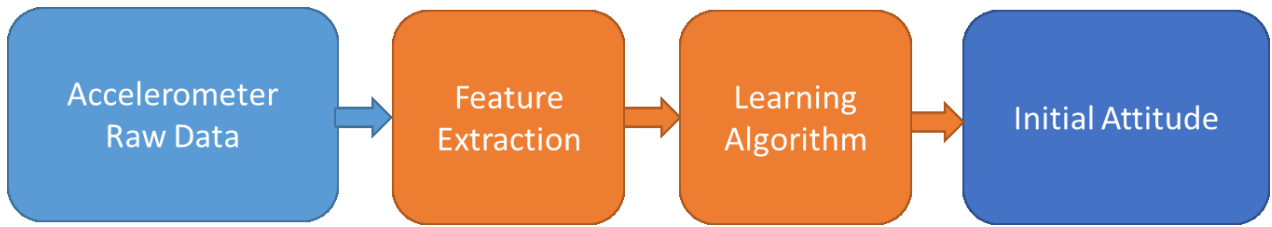

Figure 10. MLCA structure showing how the initial attitude is obtained using accelerometer readings and a learning algorithm

Ensuring effective operation of inertial sensors, regardless of the application, requires the application of several operations, including inertial sensor denoising and calibration. Further, when working with inertial sensor readings in an inertial navigation system, sometimes the initial attitude is determined by the inertial sensor readings in a process known as self-coarse alignment. All three tasks are done using model-based approaches.

*6.2 Attitude and Heading Reference System*

Attitude determination is a basic requirement in many disciplines and platforms such as motion tracking, pedestrian dead reckoning, and aerial vehicle tracking. Solving the platform orientation (commonly with inertial sensors) is done using attitude and heading reference system (AHRS) algorithms. In the literature numerous model-based AHRS approaches exist. The main challenge in AHRS is to deal with external accelerations and magnetic disturbances; for example, a walking user holding a smartphone in an indoor environment.

Our initial work in the field proposed a data driven, adaptive complementary filter to estimate the attitude (roll and pitch angles) [32]. Accelerometer weights are tuned, based on the residual of the specific force and estimated gravity vector. Such mapping is done using an empirical function modified for optimal performance by comparing to ground truth (GT) data. When our approach was evaluated and compared to commonly used AHRS algorithms including some based on complement filtering and Kalman filtering, we showed experimentally that even in presence of linear accelerations, our approach performed better in all examined scenarios by an average of 28%. Later, in our follow-up paper [33], we elaborated upon our approach to include magnetometer readings that also estimate the heading angle. We used a different experimental dataset including both linear acceleration and magnetic disturbances to show our approach outperformed all other AHRS approaches by 32%.

Recently, based on the foundation of our prior studies [32], [33], we proposed a hybrid model-learning framework for attitude estimation [34]. The learning algorithm aims to cope with the varying amplitudes caused by linear accelerations applied by a walking user. The output of this learning algorithm is introduced into the model equations to form an adaptive complementary filter structure. That is, instead of using constant or adaptive model-based weights, the accelerometer weights in each axis are regressed by a unique learning algorithm, as presented in Figure 11.

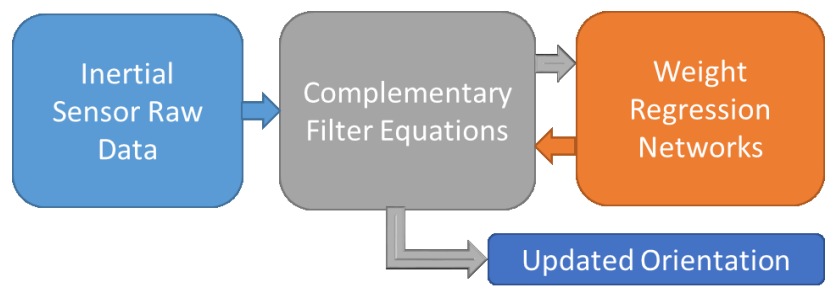

Figure 11. Hybrid structure for attitude determination represented as a recurrent neural network structure

Instead of end-to-end, deep learning solutions using a "black box" strategy (input inertial data, output attitude), the proposed method relies on a well-established model theory and uses learning methods to determine only the required weights within the model equations. Our hybrid approach managed to improve model-based approaches by approximately 40%.

*6.3 Navigation Filter*

To bind the pure inertial solution drift, the INS is commonly fused with GNSS measurements for air and land vehicles, while in the underwater environment, DVL is used in the fusion process. Commonly, the model-based extended Kalman filter in the error-state implementation is chosen to carry out the fusion process. To that end, key parameters such as the filter step size and the values of the process noise covariance matrix in the navigation filter are determined prior to its application.

The filter step size is responsible for the solution update frequency. Actually, the choice of the step size poses a tradeoff between computational load and navigation filter performance. In [35], we proposed a supervised classification learning adaptive tuning method to select the proper INS step size. We defined a bound for the velocity error allowing the navigation filter (both GNSS/INS and DVL/INS) to operate in suboptimal working conditions, and yet, in real-time, minimize the computational load. Field experiments using a quadrotor showed the advantages of our proposed approach. Also, our framework can be applied to any other navigation fusion cases between any type of sensors or platforms.

Another critical parameter of the navigation filter is the process noise covariance, usually assumed constant throughout operation. The process noise determines the filter solution accuracy, as it aims to reflect the uncertainties in the vehicle dynamics and inertial sensors. As the inertial sensor exhibits measurement variations during the vehicle motion, the value of the process noise covariance should be modified online. To that end, in the last fifty years different model-based adaptive filters were proposed. In [36], we presented a hybrid learning, adaptive navigation fusion to perform the fusion between INS and the aiding sensor. We leverage the well-established, model-based Kalman filter and construct a learning network to adaptively adjust the process noise covariance using only the inertial sensor measurements as input to the network. Our proposed approach is illustrated in Figure 12. We demonstrated the approach in a field experiment using a quadrotor and showed an improvement of 25% in the position error compared to model-based adaptive approaches.

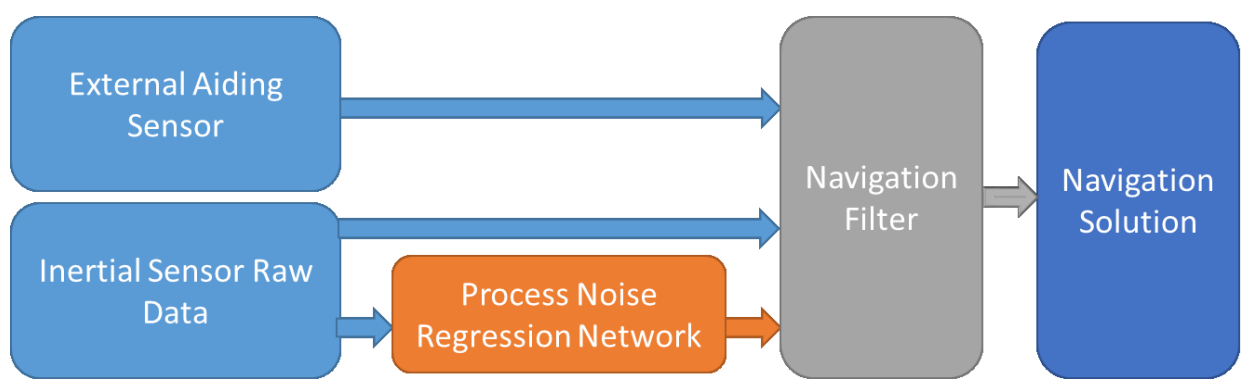

Figure 12. Our hybrid learning adaptive navigation filter block diagram

# 7 Ataxia Classification in Dogs

Ataxia is the loss of coordination of movement or the interaction of associated muscles, accompanied by a disturbance of the gait pattern. The clinical diagnosis and evaluation of its severity is usually done using subjective scales during a neurological examination. In [37] we conducted an experimental study to learn if low-cost inertial sensors, namely those available in a common smartphone, are a suitable sensor to detect ataxia. To that end, a dataset was recorded using 55 healthy dogs and 23 dogs with ataxia for a total of 770 walking sessions. In each session the accelerometer and gyroscope readings were recorded.

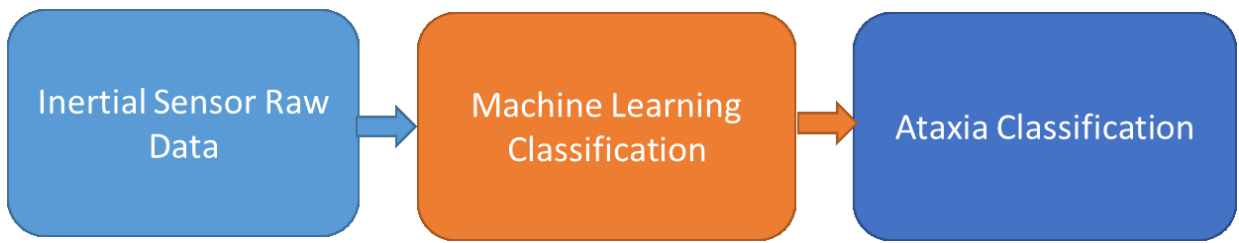

Figure 13. Raw inertial data is introduced to machine learning algorithms to classify ataxia in dogs

Several machine learning approaches were examined for the classification process using only the inertial sensor readings as input, as illustrated in Figure 13. Our results showed that the K-nearest neighbor approach obtained the best performance with 95% accuracy in discriminating between a healthy control group and ataxic dogs, indicating the potential usage of low-cost inertial sensors for canine ataxia diagnosis and monitoring of treatment effect.

# 8.    Conclusions

In this paper we reviewed 24 novel learning navigation algorithms developed at ANSFL, University of Haifa. All of those algorithms included, besides the novelty, an experimental validation of the proposed approaches. In addition, those algorithms cover a wide range of the navigation domain starting from basic navigation tasks as such inertial sensor denoising and calibration, continuing with stationary coarse alignment passing through data-driven attitude and heading reference systems, and ending with hybrid navigation filters. From an application perspective, we presented learning approaches for autonomous underwater vehicles, quadrotors, land vehicles, and indoor navigation using smartphones.

As can be observed from this vast review, we are at the beginning of exciting times where data-driven approaches meet the navigation domain to enhance current approaches' accuracy, increase their robustness, and open the door to new applications and tasks.